\documentclass{article}

\PassOptionsToPackage{numbers}{natbib}

\usepackage[preprint]{neurips_2026}

\usepackage[utf8]{inputenc} 
\usepackage[T1]{fontenc}    
\usepackage{hyperref}       
\usepackage{url}            
\usepackage{booktabs}       
\usepackage{amsfonts}       
\usepackage{nicefrac}       
\usepackage{microtype}      
\usepackage{xcolor}         
\usepackage{tikz}
\usepackage{dirtytalk}
\usepackage{natbib}
\usepackage{amsmath,amssymb}

\title{Block-Based Double Decoders}

\author{
  Vanessa Alexander\thanks{All authors contributed equally. Author names are listed alphabetically by last name. Correspondence should be addressed to Asher Labovich: \texttt{asher\_labovich@brown.edu}.} \\
  \texttt{vanessa\_alexander@brown.edu} \\
  \And
  Benjamin Bradley\footnotemark[1] \\ 
  \texttt{benjamin\_bradley@brown.edu} \\ 
  \And
  Chaitanya Harsha\footnotemark[1] \\ 
  \texttt{chaitanya\_harsha@brown.edu} \\ 
  \And
  Asher Labovich\footnotemark[1] \\ 
  \texttt{asher\_labovich@brown.edu} \\
  \AND 
  \normalfont
  Department of Computer Science\\
  Brown University\\
  Providence, RI 02912 \\ 
}

\begin{document}

\maketitle

\begin{abstract}
Encoder-decoder models offer substantial inference-time savings over decoder-only models, but their pretraining objectives suffer from sparse supervision and dynamic sequence lengths, keeping them out of practice at scale. We propose \textbf{block-based double decoders}, a novel transformer architecture that utilizes doubly-causal block-based attention masks to train with full loss supervision and static sequence packing, combining decoder-only training efficiency with encoder-decoder inference efficiency. In scaling law experiments, block-based double decoders strongly outperform encoder-decoders and closely track decoder-only models across scales. At inference time, they cut KV-cache memory and per-token compute by at least $\frac{2}{3}$ without sacrificing prefill caching or other existing inference optimizations available to decoder-only models. 
\end{abstract}

\section{Introduction}
In recent years, the rise of the transformer \citep{Attentionallyouneed} has led to advances in natural language modeling. Although the original transformer employed an encoder-decoder based architecture, with full attention for understanding in the encoder and causal attention for prediction in the decoder, decoder-only architectures have gained prominence for their scalability in text-generation settings \citep{radford2018improving}. Recently, however, there has been renewed attention in encoder-decoder architectures \citep{elfeki2025returnencodermaximizingparameter} for their gains in efficiency and their effectiveness in compute- and storage-constrained environments, as they achieve substantial reductions in KV-caching.

Even with this work, it has been shown \citep{raffel2023exploringlimitstransferlearning} that the architectural boundary between the encoder and decoder leads to inefficiencies, and that a \(2P\) parameter encoder-decoder model incurs the same computational cost as a \(P\) parameter decoder-only architecture. An alternative single-transformer approach, PrefixLM, in which full attention is used for a prefix of the inputs, and causal attention after this, attempts to combine the bidirectional context of encoder-decoders with the sharing of parameters in decoder-only models. This method benefits from less dynamic batching, but leaves many tokens untrained on.

But PrefixLM is outperformed by an encoder-decoder trained with span corruption, where the model is trained to predict a proportion of ``corrupted'' tokens. Although this achieves superior performance, it requires highly dynamic batching, and still leaves many tokens untrained on, as \citep{raffel2023exploringlimitstransferlearning} found that a mere 15\% corruption results in the best performance.

We propose a novel attention mask for pre-training\footnote{Code found at \href{https://github.com/ashlab11/block-based-double-decoder}{https://github.com/ashlab11/block-based-double-decoder}}, which we call \textbf{doubly-causal block-based masking}. Our proposed architecture consists of two decoders. The first sees a standard causal mask and takes in the standard input. The second decoder conducts cross-attention on the output of the first decoder, but also on the input, masked with our approach. We split the input into ``blocks,'' with full self-attention within each block, and causal cross-attention between blocks. In this way, we receive loss signal from every token in the input, and we achieve constant token length.

\begin{figure}
    \centering
    \includegraphics[width=1\linewidth]{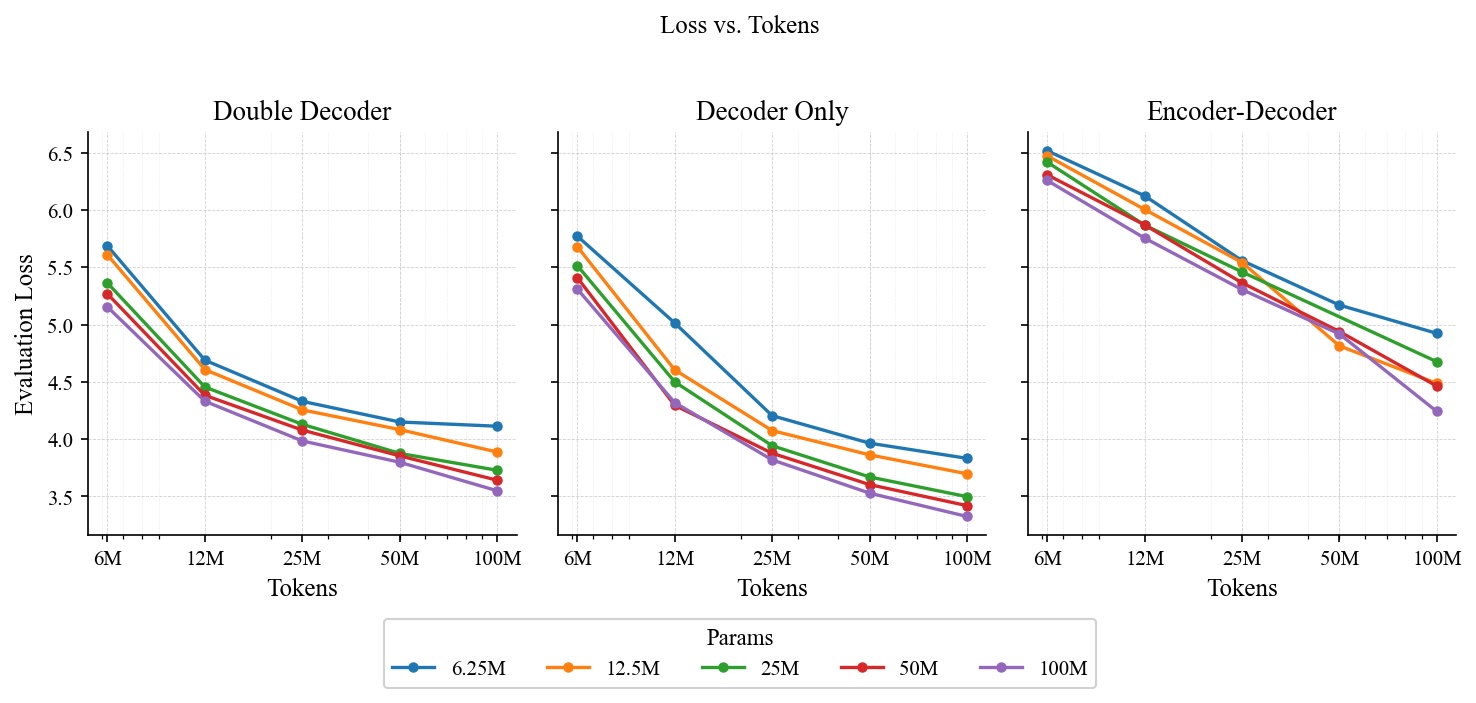}
    \caption{Graph comparing loss versus tokens for different size decoder, double decoder, and encoder-decoder models}
    \label{fig:loss_tokens}
\end{figure}

\section{Prior Work}
The original Transformer architecture \citep{Attentionallyouneed}
consists of a bidirectional self-attention encoder paired with a causal decoder for autoregressive generation. The encoder processes the full input sequence to produce contextualized representations for each token. In contrast, the decoder applies masked (causal) self-attention and additionally incorporates cross-attention, allowing each decoding position to attend to all of the encoded input representations. However, in recent years, it has been found that for generative modeling, decoder-only models are more scalable and achieve superior performance \citep{radford2018improving, brown2020languagemodelsfewshotlearners}. Meanwhile, encoder-only architectures \citep{devlin2019bertpretrainingdeepbidirectional}, trained using token masking, are performant when fine-tuned on downstream classification-type tasks.

However, \citep{raffel2023exploringlimitstransferlearning} demonstrated that any classification task can fundamentally be described as a text-to-text problem, putting it in reach of generative models. Their findings included a thorough search of several architectures and pre-training objectives, including decoder-only with the standard LM objective, decoder-only with the PrefixLM objective, and encoder-decoder with span corruption, among others. They found that across all the downstream tasks these models were tasked with, including question answering, summarizing, and translation, encoder-decoder with span corruption performed the best across a range of prediction tasks.  In span corruption, contiguous spans of tokens are "corrupted" or replaced with sentinel tokens that are assigned a unique ID specific to that sentence. The target sequence consists of the concatenated missing spans, and the model is then trained to autoregressively generate those missing spans. Typically, only a fraction of the tokens (about 15\%) are corrupted with major performance degradation occurring at 50\% corruption.

Empirically, this architecture combined with span corruption has been shown to outperform both decoder-only language models (LM) and prefix variants under comparable training settings. Additionally, span corruption as a pretraining objective has been shown to empirically outperform standard LM pretraining objectives under comparable training settings within the same model architectures.

We note that, despite its high performance, that there are several issues with span corruption as a pre-training objective. First, \citep{raffel2023exploringlimitstransferlearning} studied what rate of span corruption would be most performant, and confirmed \citep{devlin2019bertpretrainingdeepbidirectional}'s findings that 15\% span corruption led to optimal training. However, this means that only these 15\% of tokens yield a loss signal, so most tokens are not being trained on. This creates a fundamental tradeoff between increasing prediction difficulty to enhance long-range reasoning and maximizing supervision density relative to the total computational cost. Additionally, span corruption requires dynamic batching, as for a given batch, the number of tokens predicted will not be constant across all training examples. This requires token padding, a waste of compute and memory. We hypothesize that an alternate pre-training objective can perform as well as span corruption, but without these inefficiencies.

Additionally, prior work such as \citep{zaheer2021bigbirdtransformerslonger} demonstrated that attention patterns that capture both local dependencies and global information can achieve strong performance at reduced computational cost. Although we strive for train-time efficiency over $O(n)$ attention, we take inspiration from this notion with an architecture that captures both local and global interdependencies.

\section{Architecture Comparison}

\subsection{Block-based double decoders}\label{sec:arch}

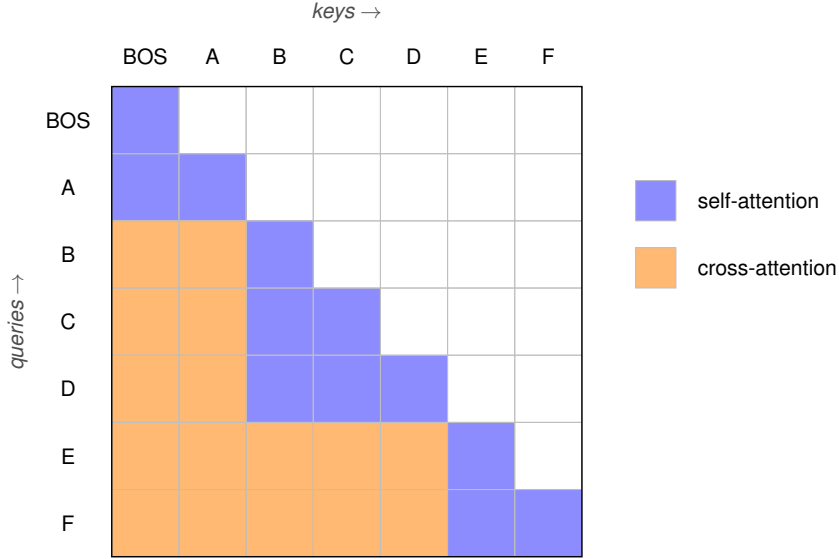
\begin{figure}[t!]
    \hspace*{2.4cm}
    \resizebox{0.8\textwidth}{!}{%
    \begin{tikzpicture}[
    scale=1,
    line width=0.4pt,
    cell/.style={draw=black!25, line width=0.3pt},
    cross/.style={fill=orange!55},
    self/.style ={fill=blue!45},
    lbl/.style  ={font=\small\sffamily},
    axlbl/.style={font=\small\sffamily\itshape, text=black!70},
]

\foreach \x in {0,...,6} {
    \fill[self] (\x, 6 - \x) rectangle ++(1, 1);
}
\fill[self] (0, 5) rectangle ++(1, 1);
\fill[self] (2, 2) rectangle ++(1, 1);
\fill[self] (2, 3) rectangle ++(1, 1);
\fill[self] (3, 2) rectangle ++(1, 1);
\fill[self] (5, 0) rectangle ++(1, 1);

\foreach \y in {0,...,4} {
  \foreach \x in {0,1} {
  \fill[cross] (\x, \y) rectangle ++(1, 1);
}
}

\foreach \x in {2,3,4} {
    \foreach \y in {0,1} {
    \fill[cross] (\x, \y) rectangle ++(1, 1);
    }
}

  \foreach \x in {0,...,6} {
    \foreach \y in {0,...,6} {
      \draw[cell] (\x,\y) rectangle ++(1,1);
    }
  }
  \draw[line width=0.6pt] (0,0) rectangle (7,7);

  \foreach \i/\L in {0/BOS, 1/A, 2/B, 3/C, 4/D, 5/E, 6/F} {
    \node[lbl] at (\i+0.5, 7.45) {\L};              
    \node[lbl] at (-0.65, 6.5-\i)  {\L};            
  }

  \node[axlbl] at (3.5, 8.1) {keys $\rightarrow$};
  \node[axlbl, rotate=90] at (-1.4, 3.5) {queries $\rightarrow$};

  \begin{scope}[shift={(7.8, 5)}]
    \fill[self]  (0,0)   rectangle (0.6,0.6);
    \draw[cell]  (0,0)   rectangle (0.6,0.6);
    \node[lbl, anchor=west] at (0.8, 0.3) {self-attention};

    \fill[cross] (0,-1)  rectangle (0.6,-0.4);
    \draw[cell]  (0,-1)  rectangle (0.6,-0.4);
    \node[lbl, anchor=west] at (0.8,-0.7) {cross-attention};
  \end{scope}

\end{tikzpicture}%
}
    \caption{Visual explanation of the decoder attention mask for an example sentence. Splitting up into three parts, we see that the decoder sees three context-response pairs in parallel: \\
    \textbf{(1):} empty context, [BOS, A] response \\
    \textbf{(2):} [BOS, A] context, [B, C, D] response \\
    \textbf{(3):} [BOS, A, B, C, D] context, [E, F] response \\
    Each token thus appears in the loss exactly once per forward pass.}
    \label{fig:double-decoder-mask}
\end{figure}%

In this paper, we wish to create an architecture for next-token prediction that retains the benefits of encoder-decoder architectures -- in particular, substantial KV-cache reductions and ease of use on edge device -- without sacrificing the training efficiency of decoder-only transformers. We consider two criteria when creating such an architecture. First, \emph{loss-information density:} every token in a packed training example should contribute a loss signal in every forward pass. Second, \textit{sequence length statis:} post-packing sequence length should not change across training batches, so that throughput is predictable and maximally efficient.

To motivate our architecture, we first consider a basic PrefixLM encoder-decoder training objective, which fails both criteria. To train one, we must randomly choose split-points at each batch, placing $N\%$ in the encoder and $(100 - N)\%$ in the decoder. Only the decoder portion contributes to the loss, so loss-information density is $(100 - N)\%$, and token counts fluctuate by batch depending on N. This results in both dynamic length batching (since N changes by batch) and lower loss information per token, which combined substantially harms training efficiency. We solve both of these problems via our proposed architecture, block-based double decoders.

Concretely, a block-based double decoder consists of two stacks. The first, which we call the \textit{context decoder}, is a standard causal decoder-only transformer; it takes in the full input and outputs causal latents $h_t$ for each token. The second, the \textit{generation decoder}, takes in three inputs: (1) the causal latents from the context decoder, (2) the token sequence itself, and (3) a \textit{block partition}: a strictly increasing sequence of indices $0 = b_0 < b_1 < \ldots < b_K = T$ that splits the sequence into K contiguous subsequences. For the length-7 sequence in Figure \ref{fig:double-decoder-mask}, the partition (0, 2, 5, 7) creates blocks [BOS, A], [B, C, D], [E, F]. 

Within the generation decoder, each query at position $t$ in block $k$ attends to two subsequences: the \textit{within-block keys}, comprising the tokens of block $k$ at positions $\le t$ (i.e. causal self-attention), and \textit{cross-block keys}, comprising the context-decoder latents $h_s$ for all $s$ in blocks $m < k$ (full cross-attention). The mask in Figure \ref{fig:double-decoder-mask} is the union of these two attention mechanisms. 

There exists a subtletly in how these two operations are combined. Treating these as two separate attention mechanisms and residually adding the outputs, as is common in normal encoder-decoders, introduces three problems. First, the order in which the two operations are applied becomes architecturally significant despite being completely arbitrary. Second, certain query rows have no keys under the cross-attention mask (e.g. [BOS, A] in Figure \ref{fig:double-decoder-mask}), leaving the corresponding softmax undefined. Third, the attention-sink behavior of standard attention is lost: with two separate softmaxes, the operation is forced to allocate equal probability mass to each of the two subsequences, even when only one is important for predicting the next token. These three problems can all be solved at once by conducting just one attention with two different key matrices depending on the (query, key) index pair. However, since to our knowledge there currently exists no fast attention implementation that allows this dual-key mechanism, we instead compute the attentions \textit{separately} and combine their log-sum-exp normalizes post-hoc. This is mathematically equivalent to conducting the singular attention operation, but allows for fast implementation via PyTorch's FlexAttention function. We note that this is substantially less efficient from a computational perspective than the \say{ideal} method, which would directly utilize the sparsity induced by the chosen blocks to minimize total multiplications across the dual SDPA applications, cutting off all additional compute beyond the calculation of an additional KV matrix. We leave the creation of this ideal method to future work.

This architecture achieves both of the criteria mentioned earlier. Every token appears in the loss exactly once per forward pass, regardless of the number of blocks chosen (though, a small first block has an empty context-decoder, and thus must conduct all reasoning in the generation-decoder). In addition, the only dynamic component is the block list, which changes each batch; however, it affects only the attention mask of the generation-decoder, which can be handled with minimal latency by PyTorch's FlexAttention. And, the existence of a context-decoder retains the benefits of normal encoder-decoder architectures: at inference-time, the context-decoder runs once over the prompt, thus requiring no KV-cache and achieving substantial speed-ups.

Each of the three architectures mentioned in this paper -- decoder-only, encoder-decoder, and double-decoder -- have substantially different compute requirements during both training and inference, even when parameter and token-matched. The following sections describe these differences in detail.

\subsection{Training Time Comparisons}\label{sec:training-time}

Although our primary focus is on the large inference-latency differences between decoder-only, encoder-decoder, and double decoder models, the three architectures also differ in training compute, and the standard $6NT$ heuristic obscures these differences. In Appendix \ref{app:training-time}, we derive architecture-aware FLOP formulas: double decoders incur additional compute from their extra KV projections, while encoder-decoders save compute by feeding fewer tokens through the decoder under span corruption. Specifically, for a decoder-only model with sequence length $T$, hidden dimension $d$, and $L$ layers, the approximate train-time FLOP count is \[L(72Td^2 + 12T^2d).\] For an encoder-decoder trained with span corruption, with padded encoder input sequence length \(T_{in}\) and decoder sequence length \(T_{out},\) this is \[L((52T_{in}+28T_{out})d^2 + (4T_{out}^2 + 4T_{in}T_{out} + 8T_{in}^2)d).\] and for an efficient implementation of the double decoder, \[L(76Td^2 + 12T^2d).\] 

Although double decoder models require more training compute than decoder-only models, the difference is small when implemented efficiently: at \(T=T_{in}=2048\) and \(T_{out}=256,\) double decoder uses only \(2.4\%\) more FLOPs than decoder-only (while encoder-decoder uses \(21\%\) \emph{fewer} FLOPs). Combining these formulas with our empirical scaling laws lets us compare the compute required by each architecture to reach matched perplexity on held-out test sets.

\subsection{Inference Time Comparison}\label{sec:inference-time}

As mentioned in Section \ref{sec:training-time}, double decoders tend to slightly \textit{underperform} decoder-only models in training compute due to their additional KV matrices in every transformer block. However, this disadvantage is strongly outweighed by their numerous benefits during inference, which we detail below. All benefits arise directly from the natural separation of context and response in both training and inference, and one such benefit is unavailable for classic encoder-decoder models due to their bidirectional nature.

\begin{enumerate}
    \item Since the generation decoder need only refer to the final output of the context decoder, there is no need to save the activations from the context decoder. This saves both time and memory; \citep{elfeki2025returnencodermaximizingparameter} found 4.7x higher throughput at low context lengths, and \citep{sun2024cacheoncedecoderdecoderarchitectures} noted these throughput gaps only increase as context length grows. In addition, the KV-cache required during generation scales only with the size of the generation decoder, rather than the full model. As detailed in Appendix \ref{app:kv-cache}, this results in a $\mathbf{\frac{2}{3}}$ KV-cache memory reduction in comparison to decoder-only models under our architectural design, with the benefit growing as the ratio of context-decoder to generation-decoder layers grows. Furthermore, these benefits carry over to per-token latency; as generation depends only on the size of the generation-decoder after the context-decoder representation is completed, the latency cost per-token scales equivalently to that of the memory boost. These advantages open up a connection to test-time compute scaling: if it is possible to inflate the compute spent on \textbf{only} the context-decoder (i.e. without adding more tokens or increasing compute in the generation-decoder) at test-time, then TTFT increases while per-token latency and memory requirements remain the same. Recently, works on looped transformers \citep{labovich2026stabilitygeneralizationloopedtransformers, prairie2026parcaescalinglawsstable, geiping2025scalingtesttimecomputelatent, saunshi2025reasoninglatentthoughtspower} have found that looping layers at test-time can improve performance and generalization capacity; doing so just for the context decoder in our work may effectively marry the positive effects on reasoning with the low memory-and-compute costs of dual-stack models. 
    \item In scenarios where memory is particularly scarce, models with two separate stacks (including both regular encoder-decoders and our block-based double decoder) can \say{swap} between CPU and GPU when needed. A model split 2/3 encoder and 1/3 decoder can save 1/3 parameter memory while applying the context decoder, before swapping the placement of each stack when generation is needed. 
    \item In cases where model providers expect many prompts to share a common prefix (e.g. a long shared system prompt) they often \textit{prefill} the KV-cache for that prefix once and reuse it across requests, dramatically reducing TTFT. While immensely useful in decoder-only models -- especially given that system prompts are typically long and thus the dominant contributor to TTFT without prefill -- this form of caching is fundamentally incompatible with encoder-decoder models: the bidirectional encoder makes every token's representation depend on the full input, so a cached prefix cannot be reused once the suffix changes. Our double decoder directly resolves this, as its doubly-causal nature allows immediate transfer of prefix-level KV caching from decoder-only inference stacks. 
    \item The architectural separation in our model also reduces latency. Time-to-first-token (TTFT) depends only on the depth of the context decoder, yielding a reduction proportional to $\frac{L_{enc}}{L}$ which corresponds to a $\mathbf{\frac{1}{3}}$ \textbf{reduction in TTFT latency} under the same split. Notably, the benefits of prefill (as used in \cite{kwon2023efficient} and other inference packages) transfer immediately over to our model in a way they \textbf{do not} to traditional encoder-decoder models due to the bidirectional nature. Given that many recent latency improvements come from improving prefill caching, this is an \textbf{immediate shift with essentially zero fundamental inference code changes.} In addition to lower TTFT, our model also substantially reduces the latency of each token generated. Per-token generation cost depends only on the depth of the generation decoder, as each new token is processed autoregressively through this stack. As a result, the latency cost per-token scales proportionally to $\frac{L_{dec}}{L}$, yielding a corresponding reduction under the same 2/3 1/3 split.
\end{enumerate}

\section{Experimental Setup}\label{sec:experimental-setup}
\paragraph{Models and scaling grid. } 
We compare three architectures -- decoder-only, standard encoder-decoder (SED), and our proposed double decoder -- all under a shared tokenizer, sequence length 2048, and tied input/output embeddings, with parameter counts approximately matched across families. The main sweep is a three-way grid over architecture, model size, and token budget, spanning 6.25M-100M parameters and 62.5M-1B tokens in multiples of 2. We vary hidden dimension across the grid with num heads = $\frac{d}{64}$ so head dimension stays at 64, holding the depth profile at 8 encoder / 4 decoder layers (or 12 decoder layers for decoder-only models) for most configurations. The smallest models use 10/5 instead, because the multiple-of-64 width constraint leaves too little resolution to hit small parameter targets by width alone. SED's decoder carries an extra cross-attention sublayer per block, so we reduce its decoder layer count to match non-embedding parameters at fixed width. We provide a detailed description of the hyperparameters constant to all models in Table \ref{tab:hyperparams}. We train all models on 10xH100 NVIDIA GPUs, with a total of 200 GPU-hours for given results. All experiments were conducted with PyTorch \cite{DBLP:journals/corr/abs-1912-01703}.

As described in Section \ref{sec:training-time}, we report architecture-aware FLOP counts rather than the common $6NT$ heuristic used for decoder-only models. 

\paragraph{$\mu$P and width transfer.}  We adopt a width-scaling protocol inspired by maximal update parameterization \citep{yang2022tensorprogramsvtuning} with base width $d_0 = 64$. Hidden-matrix learning rates are scaled by $\frac{d_0}{d}$, embeddings and the tied output projection use the architecture-specific base learning rate, and norms and biases skip weight decay. After the tied output projection, logits are multiplied by $\sqrt{\frac{d_0}{d}}$. Base learning rates are tuned on a 0.5M model for all three architectures (0.01 for decoder-only and double decoder, 0.004 for encoder-decoder) and confirmed to scale to models 16x at large. After confirming the ideal learning rate, we conducted a sweep of weight decay along the 0.5M param model, and found 0.5 best for double decoder and encoder decoder with 0.1 best for decoder-only. Appendix \ref{app:muP} provides a visual description of these sweeps.

\paragraph{Pretraining data and objectives.} All models pretrain on packed SlimPajama \citep{cerebras2023slimpajama} with architecture-native objectives: double decoder uses our doubly-causal block-based masking method, SED uses T5-style span corruption with sentinels, and decoder-only uses causal next-token prediction. Because these losses correspond to different prediction problems, we do not compare raw pretraining losses across families. Instead, we add a post-hoc prefix-LM fine-tuning phase (10\% of training data) on held-out SlimPajama with a base learning rate of 0.0002 before applying $\mu$P.  

The collator samples a breakpoint at each batch and trains the model to predict suffix tokens conditioned on prefix tokens. For double decoder and SED, the prefix is routed through the encoder; for decoder-only, the prefix merely does not provide additional loss information. This equalizes the loss objective and data and allows for stronger cross-architecture comparisons.

\section{Results}
\begin{figure}
    \centering
    \includegraphics[width=1\linewidth]{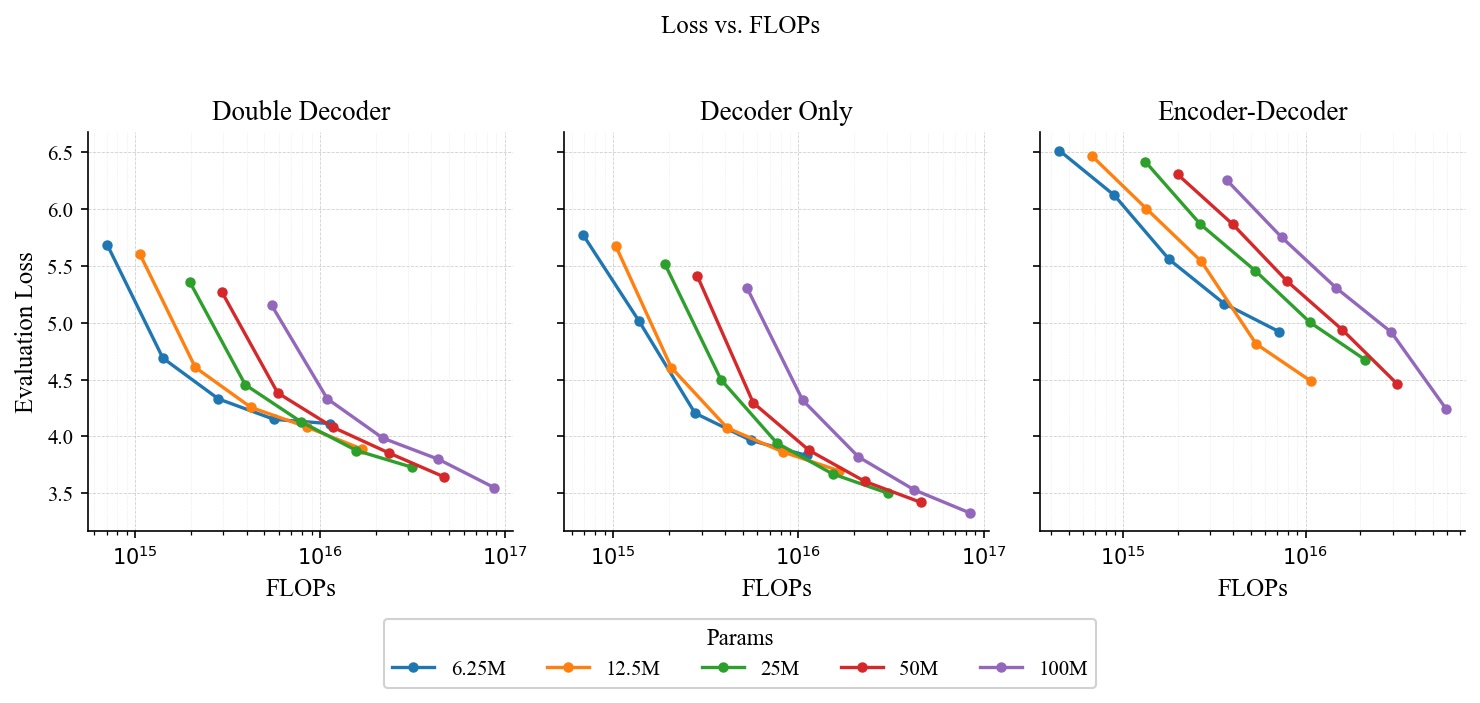}
    \caption{Graph of loss vs FLOPs curve for each size model}
    \label{fig:loss_flops}
\end{figure}

\begin{figure}
    \centering
    \includegraphics[width=1\linewidth]{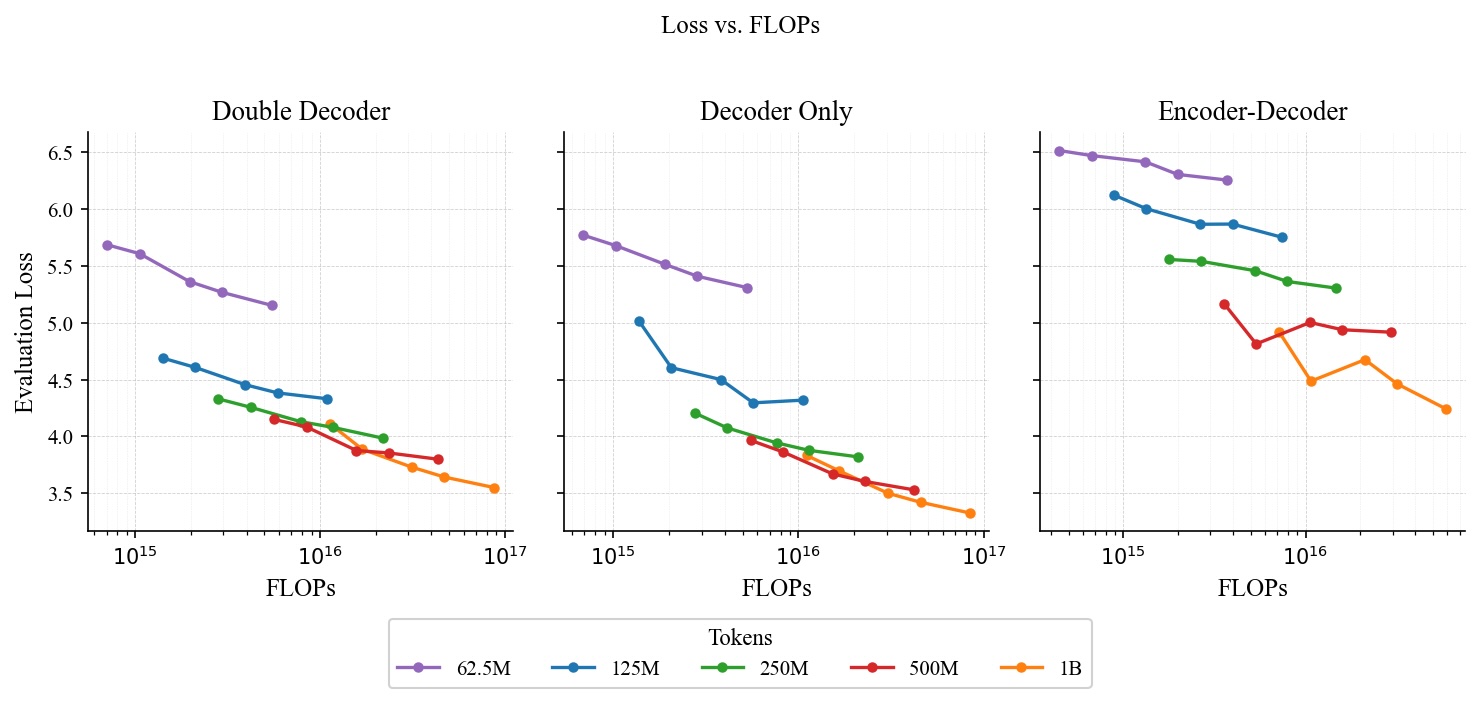}
    \caption{Graph of loss vs FLOPs curve for models by token training count}
    \label{fig:loss_flops_tokens}
\end{figure}

Across the entire (N, D) grid, our block-based double decoder substantially outperforms the encoder-decoder baseline and tracks decoder-only closely (Figures \ref{fig:loss_tokens}, \ref{fig:loss_flops}, \ref{fig:loss_flops_tokens}). At the largest configuration we evaluated (100M params, 1B tokens), the double decoder reaches an evaluation loss $\sim 0.2$ nats worse than the parameter and token-matched decoder-only model, while the encoder-decoder trails \textit{both} architectures by $\sim 0.7$ nats. This relative ordering -- decoder-only $\lesssim$ double decoder $\ll$ encoder-decoder -- holds across every model size and token budget in the sweep. 

The most informative difference is visible along the compute axis in Figure \ref{fig:loss_flops}. For both decoder-only and double decoder, individual size-curves cross over, tracing out the Pareto envelope characteristic of well-known scaling laws \citep{hoffmann2022trainingcomputeoptimallargelanguage}. Yet, across the entire FLOP range tested, larger encoder-decoder models are essentially \textit{uniformly} worse than smaller ones at every fixed compute level, and the curves remain nearly parallel on the log axis. We interpret this as direct evidence that all tested encoder-decoder configurations sit in the same scaling regime -- specifically, the \textbf{data-limited} regime, where additional parameters cannot be utilized within the token budgets explored. Figure \ref{fig:loss_tokens} provides additional evidence for this claim: while the decoder-only and double decoder curves flatten visibly past $\sim 50$M tokens as smaller models hit their capacity floors, the encoder-decoder curves continue descending steeply at every model size with no sign of flattening. This finding directly motivates and justifies block-based double decoders. As argued earlier, span corruption only receives loss information from $\sim 15\%$ of tokens during the forward pass, so the encoder-decoder receives a sparse signal and cannot fill its capacity within the token budgets we test. By restoring full token loss information via block-based double decoders, we immediately see substantial gaps across the entire compute grid.

\begin{figure}
    \centering
    \includegraphics[width=1\linewidth]{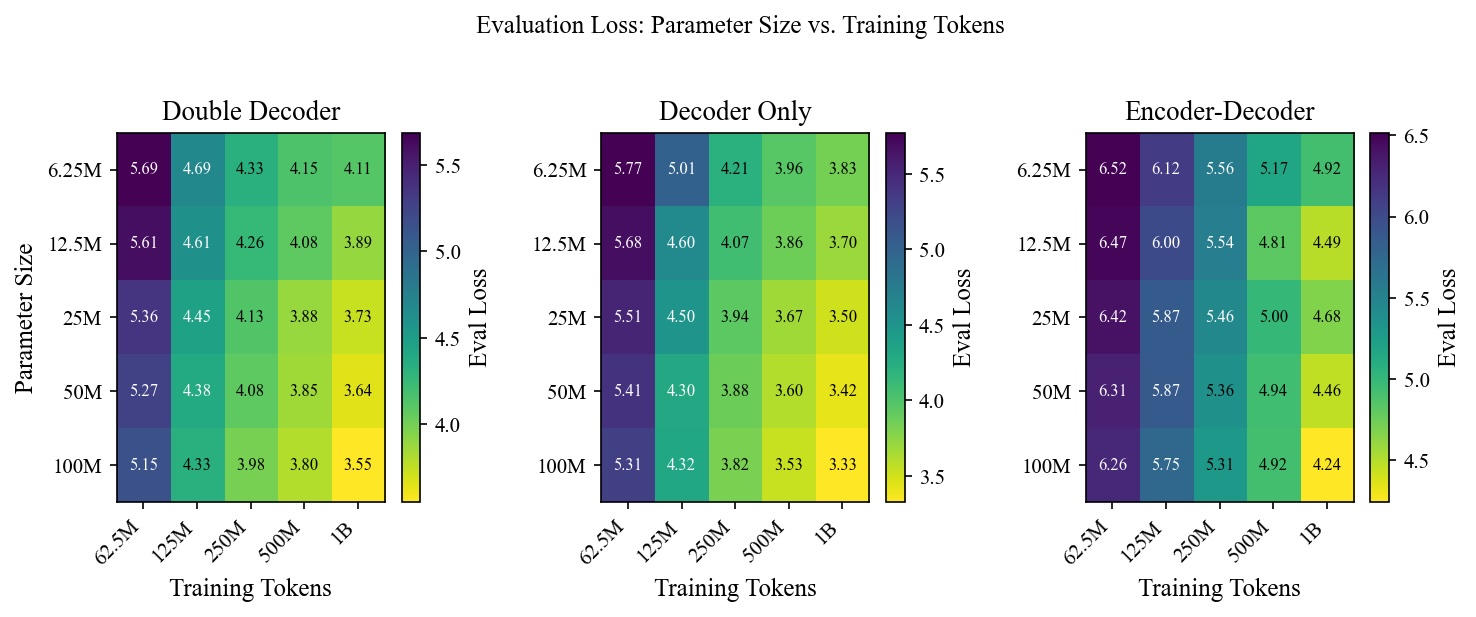}
    \caption{Table showcasing CE loss after training for each parameter/token combination.}
    \label{fig:table}
\end{figure}

The $\sim 0.2$ nat gap between the decoder-only model and the double decoder model is what we believe to be the cost of the architectural separation of context and response. Section \ref{sec:inference-time} details the inference-time benefits this separation enables, which we view as more than compensating for the modest training-time loss gap. 

\section{Limitations and Future Work}
While our method achieves strong results, we note several limitations. First, attention blocks are sampled randomly per batch, which may increase variability and unpredictability of loss dynamics. Second, due to compute constraints we were limited to relatively small-scale models, and our scaling-law extrapolations may not hold at several orders of magnitude beyond the regime we directly consider. Training a select few larger models would substantially strengthen evidence for extrapolation at much larger scales.

Beyond the inference time savings mentioned here, the double decoder architecture opens several promising research directions. Latent chain-of-thought methods such as COCONUT
\citep{hao2024training} find benefits in inserting continuous reasoning tokens between context and response, but the absence of a native context-response boundary in decoder-only architectures has confined these techniques to post-training. Our architecture exposes this separation throughout the entire training process, theoretically opening the door for latent reasoning to be applied during pretraining. As discussed in Section \ref{sec:inference-time}, looped transformers \citep{labovich2026stabilitygeneralizationloopedtransformers, prairie2026parcaescalinglawsstable} also naturally synergize with the context decoder, possibly combining test-time reasoning gains with low-latency inference. We leave direct testing of these possibilities to future work.

\newpage

\bibliographystyle{unsrtnat}
\bibliography{bibliography}
\newpage

\appendix
\section{Additional Calculations and Results}
\subsection{KV-cache calculations at inference time}\label{app:kv-cache}
Here, we calculate the difference between a decoder-only and dual-stack (encoder-decoder, or double-decoder) model in terms of key architectural parameters. We define the following variables:
\begin{itemize}
    \item $d:$ hidden state size
    \item $L_{enc}$: layers in encoder / context-decoder
    \item $L_{dec}$: layers in decoder (as part of encoder-decoder) / generation-decoder
    \item $L$: layers in decoder-only model
    \item $T_{in}$: number of tokens in context
    \item $T_{out}$: number of tokens in output
    \item $b$: bytes per activation (e.g. 2 in fp16)
\end{itemize}
We assume MHA for attention and KV-caching for simplicity. 

For a decoder-only model, calculating the memory of the KV-cache is simple: every token has both a key and value representation for every block. Thus, memory requirement can be calculated as \[
2dbL(T_{in} + T_{out})
\]
For dual-stack models, calculating is more complicated. For self-attention, each token in $T_{out}$ attends only to prior tokens in $T_{out}$ across all decoder-layers, giving a cache of $2dbL_{dec}T_{out}$. For cross-attention, each layer has a key and value representation of the final encoder state; thus, the cache is $2dbL_{dec}T_{in}$. Combined, this results in a cache of \[
2dbL_{dec}(T_{in} + T_{out})
\]
As the only difference at inference-time between double decoders and encoder-decoders is the use of log-sum-exp rather than summing the attention blocks together, both possess the same memory advantages. In particular, the memory cost in KV-cache relative to a decoder-only model can be represented simply as \[
\frac{L_{\text{dec}}}{L}
\]
Since our models utilize a 2/3 1/3 dual-stack split, this results in a $\mathbf{\frac{2}{3}}$ KV-cache memory reduction.
\subsection{FLOP heuristic calculations}\label{app:training-time}
With variables as defined in the last section, we work through a single head in a single layer of self attention. In a decoder only, there are 4 linear matrix multiplications, so \(8Td^2,\) two matrix multiplications for SDPA, \(4T^2d,\) and the two FF layers contribute \(16Td^2.\) Together, this is \[24Td^2+4T^2d.\] We multiply by \(L\) for the number of layers, and since we are studying FLOPs at train time, we multiply by 3, for the forward pass, and then both backwards passes (for weights and for activations). Hence \[L(72Td^2+12T^2d).\] The formula for encoder-decoder is calculated in a similar way. Since our models utilize a \(2/3\) \(1/3\) split, we multiply by \(\frac{2L}{3}\) for the encoder and \(\frac{L}{3}\) for the decoder. The FLOP count for a single encoder layer is identical to that of a single decoder, so \[3\cdot\frac{2}{3}L(24T_{in}d^2+4T_{in}^2d) = L(48T_{in}d^2+8T_{in}^2d).\] For the decoder, the dual cross-attention on the encoder's output and self-attention on the smaller token size complicates the calculation somewhat, but we find \[L(28T_{out}d^2+4T_{out}^2d+4T_{in}T_{out}d+4T_{in}d).\] Summing these two yields \[L((52T_{in}+28T_{out})d^2 + (4T_{out}^2 + 4T_{in}T_{out} + 8T_{in}^2)d).\] Finally, for the double decoder architecture, the context-decoder is a standard causal decoder, so the FLOP count is once again \[L(48Td^2+8T^2d).\] The generation-decoder has six linear matrix multiplications and 2 SDPA modules, at least in our naive implementation. However, a theoretically optimal matrix multiplication algorithm would bring the computation down to the equivalent of 1 SDPA, because both decoders have a causal mask, so the necessary multiplications of query, key, and value matrices are equivalent to an unmasked attention head. Hence, the linear matmuls yield \(12Td^2,\) the SDPA yields \(4T^2d,\) and the two FF layers add \(16Td^2.\) Hence, the generation-decoder uses \[L(28Td^2+4T^2d),\] altogether \[L(76Td^2+12T^2d).\]

\subsection{$\mu$P hyperparameter sweeps}\label{app:muP}
The following graphs showcase our findings for sweeps on learning rate and weight decay using $\mu$P.

\begin{figure}[h!]\label{fig:muP}
    \centering
    \includegraphics[width=\linewidth]{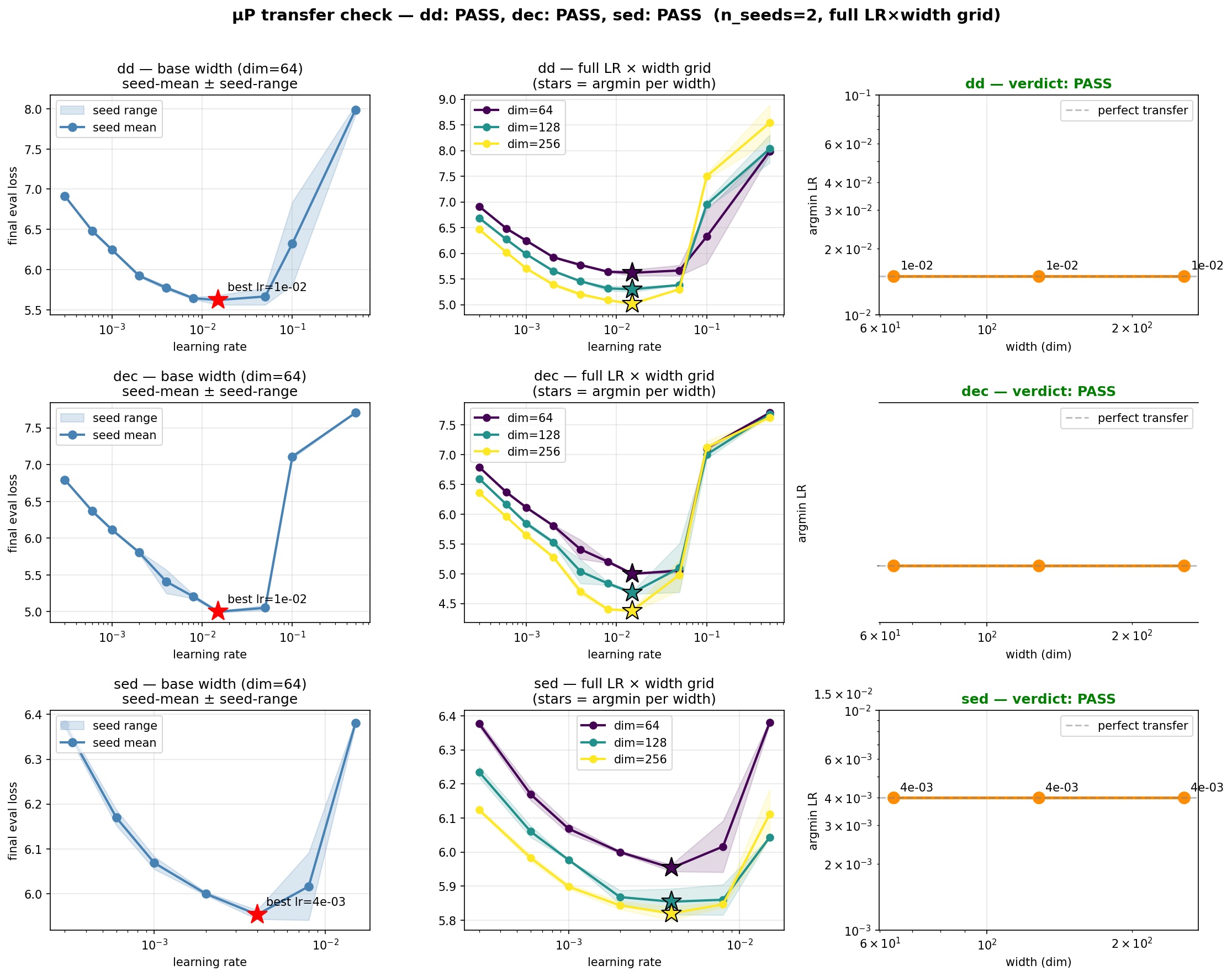}
    \caption{Graphs showcasing the results of $mu$P learning rate transfer. Left graphs find the best learning rate at the smallest model (0.5M). Middle graphs compare that learning rate with others for larger models. Right graphs confirm that the best learning rate remains constant across scales.}
    \label{fig:placeholder}
\end{figure}

\begin{figure}[h!]\label{fig:weight-decay}
    \centering
    \includegraphics[width=\linewidth]{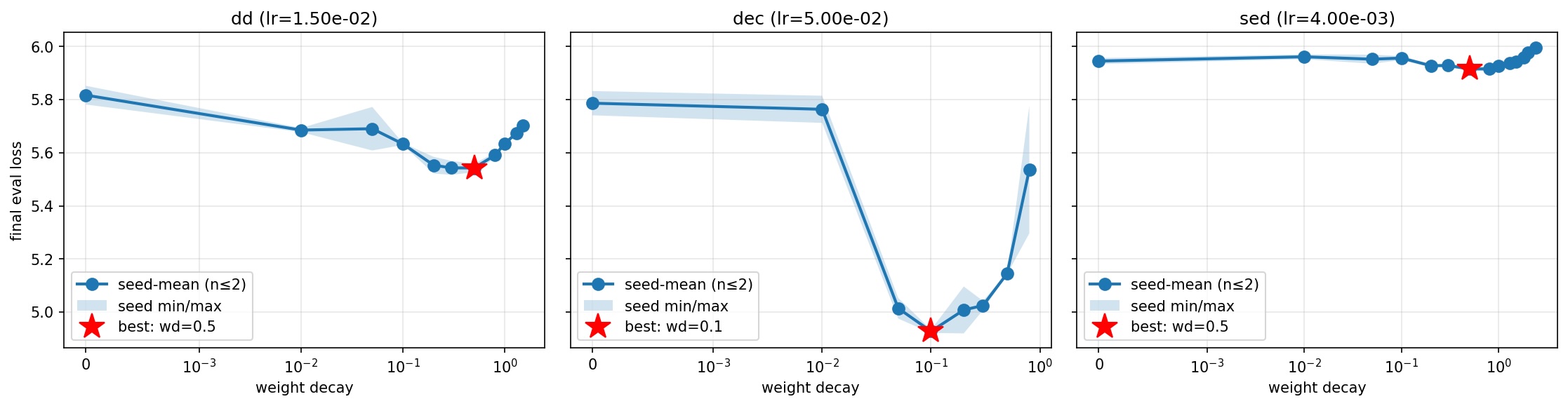}
    \caption{Graphs showcasing the result of weight decay sweeps after finding the ideal learning rate.}
    \label{fig:placeholder}
\end{figure}

\subsection{Common hyperparameter table}
\begin{table}[t]
\centering
\small
\begin{tabular}{lll}
\hline
\textbf{Category} & \textbf{Hyperparameter} & \textbf{Value} \\
\hline
Data & Sequence length & 2048 tokens \\
Data & Tokenizer & 32k BPE tokenizer \\
Initialization & Weight initialization & Xavier uniform \\
Architecture & Input/output embeddings & Tied \\
Architecture & Attention head dimension & 64 \\
Architecture & Positional encoding & Rotary positional embeddings \\
Optimization & Optimizer & AdamW \\
Optimization & AdamW betas & $(0.9, 0.95)$ \\
Optimization & AdamW epsilon & $10^{-8}$ \\
Optimization & LR schedule & Linear warmup for 5\% of steps, then linear decay \\
Optimization & Final LR fraction & 0.1 of peak LR \\
Optimization & Gradient clipping & Global norm clipped to 1.0 \\
$\mu$P & Base width & 64 \\
$\mu$P & Hidden-weight LR rule & $\eta_{\mathrm{hidden}} = \eta_{\mathrm{base}} \cdot 64 / d$ \\
$\mu$P & Embedding/output/norm LR & $\eta_{\mathrm{base}}$ \\
Pretraining & Base LR source & Per-architecture tuned LR, unless overridden \\
Pretraining & Weight decay source & Per-architecture tuned WD, unless overridden \\
PrefixLM SFT & Base LR & $2 \times 10^{-4}$ \\
PrefixLM SFT & Hidden-weight LR & $2 \times 10^{-4} \cdot 64 / d$ \\
PrefixLM SFT & SFT token budget & 10\% of pretraining tokens by default \\
PrefixLM SFT & Effective batch size & 32 sequences \\
\hline
\end{tabular}
\caption{Hyperparameters shared across model families. Here $d$ denotes model width. Architecture size, layer allocation, pretraining objective, and collator differ by model family and are therefore omitted.}
\label{tab:hyperparams}
\end{table}

\end{document}